\documentclass[runningheads]{llncs}

\usepackage{eccv}

\usepackage{eccvabbrv}

\usepackage{graphicx}
\usepackage{booktabs}

\usepackage[accsupp]{axessibility}  %

\usepackage{hyperref}

\usepackage{orcidlink}

\usepackage{catchfile} %
\usepackage{adjustbox}
\usepackage{multirow}
\usepackage{colortbl}

\usepackage[accsupp]{axessibility} %

\usepackage{lipsum}
\usepackage{array}
\usepackage{makecell}
\usepackage{soul}
\newcolumntype{H}{>{\setbox0=\hbox\bgroup}c<{\egroup}@{}}
\usepackage{color}
\usepackage{stfloats}
\usepackage{float}
\usepackage{placeins}

\definecolor{light}{rgb}{0.5, 0.5, 0.5}

\definecolor{bestblue}{rgb}{0.70, 0.84, 1.00}
\definecolor{secondblue}{rgb}{0.88, 0.94, 1.00}
\definecolor{improvegreen}{rgb}{0.86, 0.95, 0.86}
\definecolor{degradered}{rgb}{0.98, 0.86, 0.86}
\newcommand{\bestcell}[1]{\cellcolor{bestblue}#1}
\newcommand{\secondcell}[1]{\cellcolor{secondblue}#1}
\newcommand{\improvecell}[1]{\cellcolor{improvegreen}#1}
\newcommand{\degradecell}[1]{\cellcolor{degradered}#1}
\usepackage{amsmath}

\definecolor{applegreen}{rgb}{0.05, 0.60, 0.20}

\newcommand{\PAR}[1]{\vskip4pt \noindent {\bf #1~}}
\newcommand{\PARbegin}[1]{\noindent {\bf #1~}}

\usepackage{pifont}
\newcommand{\xmark}{\text{\ding{55}}}

\begin{document}

    \title{A Controlled Study of Self-Supervised Image and Video Pretraining under Limited Resources}
    \titlerunning{Image and Video SSL Pretraining under Limited Resources}

    \author{Brunó B. Englert\orcidlink{0009-0001-5614-5490}  \quad Gijs Dubbelman\orcidlink{0000-0001-6635-3245}}

    \authorrunning{B.B.~Englert and G.~Dubbelman}

    \institute{Eindhoven University of Technology \\
    \email{\{b.b.englert, g.dubbelman\}@tue.nl}}

    \maketitle

    \begin{abstract}
    Visual foundation models are a cornerstone of image and video understanding but typically require large amounts of data and computation.
    The current scale required for pretraining visual foundation models may be unsustainable or unnecessary, and significant benefits arise when effective models can be obtained with fewer resources.
    To better understand how self-supervised learning (SSL) objectives behave under resource constraints, we conduct a controlled study of image and video SSL objectives under matched data, architecture, and compute budgets.
    We compare contrastive, reconstruction, feature-prediction, and diffusion objectives and evaluate both standalone and jointly trained image-video SSL formulations across a diverse set of image and video understanding tasks.
    Our results show that DINOv2-style pretraining consistently provides the strongest overall performance under limited resources. Furthermore, combining DINOv2 with video SSL objectives such as VideoMAE substantially improves image classification and segmentation performance, but degrades video tracking and camera-pose estimation performance, revealing an important tradeoff between semantic and geometric representation learning.
    These findings suggest that combining image and video SSL objectives can be beneficial in resource-limited settings, while highlighting the need for improved methods that better balance semantic, temporal, and geometric supervision.
    The implementation is available at  \texttt{\href{https://github.com/tue-mps/vision-ssl-study}{github.com/tue-mps/vision-ssl-study}}.


\end{abstract}

    \section{Introduction}
\label{sec:intro}

Modern neural vision systems are increasingly expected to generalize across a wide range of tasks and deployment settings.
A single visual backbone may be used for classification, retrieval, dense prediction, tracking, or multimodal reasoning after deployment.
This requirement is particularly important for multimodal systems such as visual large language models, where a frozen visual encoder must provide representations that remain useful across diverse prompts, tasks, and prediction heads.
To meet these requirements, visual backbones are increasingly pretrained as \emph{visual foundation models} (VFMs), whose goal is to learn transferable visual representations rather than specialize to a single downstream task.

Self-supervised learning (SSL) is an attractive paradigm for training VFMs because it turns the visual data itself into a supervision signal.
Rather than optimizing for a fixed annotation space, SSL can exploit appearance, geometry, context, and temporal structure from large image and video collections.
As a result, SSL has become a cornerstone of modern visual representation learning and underpins many recent image and video foundation models.
However, state-of-the-art SSL-based VFMs are typically obtained using massive datasets, long training schedules, and substantial computational resources.
This raises an important practical question: can useful VFMs be obtained under significantly smaller compute budgets?
The answer matters because resource-efficient pretraining would enable rapid experimentation, retraining after deployment distribution shifts, adaptation to private or site-specific data, and broader accessibility of foundation-model development.

To understand whether effective VFMs can be obtained with fewer resources, it is necessary to first understand how different SSL objectives behave in compute-constrained settings.
Unfortunately, this remains unclear.
Existing SSL methods are often pretrained using different source datasets, model architectures, input resolutions, training schedules, and optimization budgets.
Table~\ref{tab:method_taxonomy} illustrates this lack of normalization across representative image and video SSL methods.
Consequently, reported differences between methods reflect not only their learning objectives but also differences in training conditions.
Yet under a fixed budget, the SSL objective determines how each update is spent and therefore becomes a key design choice.
A controlled comparison is therefore essential for understanding which objectives are most effective when data, model capacity, and compute are held constant.

A second open question concerns \emph{objective complementarity}.
When training resources are abundant, continued optimization of a single objective may be sufficient to obtain strong representations.
For example, MAE continues to improve with substantially longer pretraining schedules and shows no indication of saturation even after 1600 epochs~\cite{he2022mae}.
Under limited budgets, however, representation quality may be constrained by the information emphasized by a single objective.
This suggests that combining objectives that focus on different aspects of visual structure may produce stronger representations than any individual objective within the same training budget.
Despite the practical importance of this question, most large-scale SSL studies train a single objective-specific recipe or compare released checkpoints, making it difficult to determine whether complementarity persists once data and compute are matched.

The first potential source of complementarity arises from semantic and reconstruction SSL objectives.
Contrastive and self-distillation methods such as MoCo~\cite{he2020moco}, SimCLR~\cite{chen2020simclr}, BYOL~\cite{grill2020byol}, MoCo v3~\cite{chen2021mocov3}, and DINO~\cite{caron2021dino,oquab2023dinov2} demonstrate that strong semantic structure can emerge without labels.
In contrast, reconstruction-based objectives~\cite{he2022mae,wei2022maskfeat,tong2022videomae} encourage models to predict missing visual content from surrounding context.
Although these objectives are often weaker on semantic recognition benchmarks, they may encourage retention of local spatial structure that is useful for dense prediction tasks such as semantic segmentation and monocular depth estimation.
This suggests that semantic and reconstruction objectives may provide partially complementary supervision signals.
A semantically strong representation such as DINOv2 may capture object-level structure, while pixel-level reconstruction may encourage preservation of fine-grained spatial information.
This motivates the study of DINOv2 combined with pixel-level SSL objectives under a fixed pretraining budget.

A second source of complementarity may arise from combining image- and video-based SSL.
Large-scale video pretraining has been shown to produce strong image-level representations~\cite{girdhar2023omnimae,wang2024internvideo2,carreira2024scaling4d}, while many successful video systems begin with a strong image encoder and add a temporal component for video understanding~\cite{ravi2024sam2,norouzi2026videomt}.
These observations suggest that much of visual representation learning may be achievable at the image level, with video pretraining contributing information related to motion, correspondence, viewpoint variation, temporal consistency, and geometry.
Consequently, jointly optimizing image and video SSL objectives may be particularly advantageous under limited compute budgets.
Conversely, video supervision may improve image representations by exposing the encoder to temporal cues unavailable in static images.

Studying this question requires an architecture that supports both image and video SSL objectives while sharing the same visual representation.
Existing image and video SSL methods often employ objective-specific architectures, making controlled comparisons difficult.
We therefore adopt a simple and unified design consisting of a shared ViT image encoder paired with a temporal module that performs full bidirectional space--time self-attention over all tokens produced across video frames.
This design allows image objectives to supervise the shared visual backbone directly, while video objectives provide additional temporal supervision through the same encoder.

Using this framework, we compare image-only SSL, video-only SSL, and joint image--video SSL under matched source data, image encoder architecture, input resolution, clip length, and pretraining budget.
All methods are pretrained on K700~\cite{carreira2019kinetics700}, with image objectives sampling individual frames and video objectives sampling clips.
The evaluated objective set includes DINOv2, I/V-JEPA variants, I/V-MAE variants, I/V-Diffusion variants, and DINOv2-anchored joint formulations.
All downstream evaluations freeze pretrained modules and train lightweight task-specific heads.
The benchmark suite spans image classification, semantic segmentation, monocular depth estimation, action recognition, tracking, and relative camera pose estimation.

This work is motivated by three practical questions.
First, how do modern SSL objectives rank when training data, architecture, and compute budget are matched?
Second, can pixel-level SSL complement a strong semantic image objective under a fixed-resource regime?
Third, can image-based and video-based SSL objectives be combined advantageously within the same pretraining budget?

\textbf{Our contributions.}
\textbf{(1)} We perform a controlled comparison of contrastive-style, reconstruction-style, feature-prediction, and diffusion-style SSL objectives for image and video pretraining under matched K700 source data, architecture choices, and an 80k-update budget, providing a benchmark for resource-efficient VFM development.
\textbf{(2)} We study objective complementarity by combining DINOv2 with both pixel-level image objectives and video SSL objectives under a fixed pretraining budget.
\textbf{(3)} We evaluate the resulting representations across a diverse suite of image and video tasks, including classification, segmentation, depth estimation, action recognition, tracking, and relative camera pose estimation, to assess which forms of visual structure are preserved by different SSL objectives.

    \section{Related Work}
\label{sec:relwork}

\begin{table}[t]
    \caption{\textbf{Original pretraining settings are not directly comparable.}
    Representative original-paper/public-code settings show that objective comparisons in the literature mix data scale, modality, resolution, patch geometry, number of training phases, update count, batch size, model size, and EMA targets. }
    \label{tab:method_taxonomy}
    \centering
    \scriptsize
    \setlength{\tabcolsep}{2.3pt}
    \renewcommand{\arraystretch}{1.12}
    \begin{adjustbox}{width=\linewidth}
        \begin{tabular}{@{}llcccccccccc@{}}
            \toprule
            Method & Data & Size & Input & Patch & Phases & Iters & Batch & Model & EMA & Modality & SSL loss \\
            \midrule
            DINOv2~\cite{oquab2023dinov2}
            & LVD-142M
            & 142M
            & \makecell{224/98\\+518}
            & 14
            & 2
            & 635K
            & 3072
            & \makecell{ViT-g\\1.1B}
            & \checkmark
            & image
            & \makecell{contrast.\\+MIM} \\

            MAE~\cite{he2022mae}
            & IN1K
            & 1.28M
            & $224^2$
            & 16
            & 1
            & 499K
            & 4096
            & \makecell{ViT-L\\304M}
            & --
            & image
            & \makecell{pixel\\recon.} \\

            I-JEPA~\cite{assran2023ijepa}
            & IN1K
            & 1.28M
            & $224^2$
            & 14
            & 1
            & 188K
            & 2048
            & \makecell{ViT-H\\632M}
            & \checkmark
            & image
            & \makecell{feature\\pred.} \\

            VideoMAE~\cite{tong2022videomae}
            & K400
            & 240K
            & $16{\times}224^2$
            & $2{\times}16^2$
            & 1
            & 375K
            & 1024
            & \makecell{ViT-H\\633M}
            & --
            & video
            & \makecell{pixel\\recon.} \\

            V-JEPA~\cite{bardes2024vjepa}
            & VideoMix2M
            & 2.0M
            & $16{\times}224^2$
            & $2{\times}16^2$
            & 1
            & 90K
            & 3072
            & \makecell{ViT-H\\$\sim$630M}
            & \checkmark
            & video
            & \makecell{feature\\pred.} \\

            JiT~\cite{li2025jit}
            & IN1K
            & 1.28M
            & $256^2$
            & 16
            & 1
            & 751K
            & 1024
            & \makecell{JiT-H\\953M}
            & --
            & image
            & \makecell{pixel\\denoise} \\
            \bottomrule
        \end{tabular}
    \end{adjustbox}
\end{table}

\PARbegin{Contrastive-style SSL.}
Large-scale SSL ViTs are now common starting points for general-purpose visual backbones.
Contrastive-style methods learn invariances by aligning different views or assignments in feature space, and later variants show that strong representations can be learned without contrasting every image against many other images~\cite{he2020moco,chen2020simclr,grill2020byol,chen2021simsiam,zbontar2021barlow,bardes2022vicreg,chen2021mocov3}.
DINOv2 is a contrastive-style teacher--student method that combines DINO, SwAV, and iBOT-style objectives with image- and patch-level targets~\cite{caron2020swav,caron2021dino,zhou2022ibot,oquab2023dinov2}.
It uses Sinkhorn-Knopp assignments from SwAV and improves training stability with an EMA teacher.
These objectives have been especially strong for semantic understanding and dense prediction~\cite{oquab2023dinov2}.

\PAR{Masked reconstruction SSL.}
Masked reconstruction trains a model to infer missing content from visible context.
MAE established masked pixel reconstruction as a scalable image-pretraining objective for ViTs~\cite{he2022mae}.
V-MAE extends the same principle to video by masking spatiotemporal tokens, so the prediction target depends on both appearance and temporal context~\cite{tong2022videomae}.
A practical drawback is training length.
MAE reports continued gains with longer pretraining and no saturation of linear-probing accuracy at 1600 epochs~\cite{he2022mae}.
V-MAE also uses long video-pretraining schedules~\cite{tong2022videomae}.
The broader masked-prediction literature varies the reconstruction target, teacher signal, or image/video data mixture~\cite{wei2022maskfeat,bao2022beit,xie2022simmim,baevski2022data2vec,girdhar2023omnimae,wang2023mvd,hernandez2024vicmae,ahamed2025crossvideomae}.

\PAR{Feature prediction SSL.}
Feature prediction shares the masked context-to-target structure of MAE-style training, but predicts learned feature targets instead of pixels.
I-JEPA samples visible context blocks and hidden target blocks from the same image, then trains a predictor to infer the target-block representations from the context representation~\cite{assran2023ijepa}.
Because these targets are learned representations, I-JEPA and V-JEPA use an EMA target encoder to stabilize training~\cite{assran2023ijepa,bardes2024vjepa}.
The model is therefore not required to reconstruct low-level texture or color details.
V-JEPA extends the same idea to video by predicting representations of masked spatiotemporal regions from visible spatiotemporal context~\cite{bardes2024vjepa}.
This places I-JEPA/V-JEPA close to MAE/V-MAE in masking structure, while separating them by the prediction target.

\PAR{Diffusion-style SSL.}
Diffusion-style SSL is related to masked prediction but not identical.
The model observes corrupted inputs and learns to recover clean signals under different noise levels.
JiT~\cite{li2025jit} brings diffusion models to image transformers by training a plain ViT to predict clean images from noisy inputs.
Because JiT keeps a plain ViT backbone, diffusion-style SSL can be compared with DINOv2, MAE, and JEPA under the same encoder architecture.
Video diffusion extends this objective to clips, making it relevant to motion, correspondence, and geometry~\cite{velez2025fromimagetovideo,ke2024marigold}.
A practical drawback is optimization length, since diffusion-transformer training schedules commonly use hundreds of thousands to millions of updates~\cite{peebles2023scalable,li2025jit}.

\PAR{Image and video evaluation.}
Benchmarks such as Kinetics and SSV2 made video evaluation standard~\cite{carreira2017quovadis,goyal2017something}.
However, action accuracy alone does not fully characterize a video representation.
Scaling 4D representations is one of the closest studies to our setting because it compares large image and video representations across a broader evaluation suite and emphasizes that recognition, geometry, and temporal behavior can lead to different conclusions~\cite{carreira2024scaling4d}.
Its broader suite includes tasks beyond recognition, making it closer to the image, video, and geometry setting considered in this work.
This kind of broad evaluation is important because representations that look similar on classification can differ on dense prediction, tracking, pose, or other tasks that require spatial and temporal structure.
Benchmarking and diagnostic studies make a related point by showing that benchmark coverage and adaptation protocols can strongly affect the apparent ranking of representation learners~\cite{zhai2020vtab,zamir2018taskonomy,danier2025depthcues}.
At the same time, many image and video foundation-model comparisons rely on public checkpoints~\cite{girdhar2023omnimae,hernandez2024vicmae,ahamed2025crossvideomae,wang2024internvideo2}.
Their objectives, data mixtures, model sizes, training budgets, resolutions, and temporal architectures differ simultaneously.
Such studies are valuable for understanding released systems, but they do not isolate the effect of the SSL objective under matched resource constraints.

    \section{Method}
\label{sec:method}

\subsection{Preliminaries}
\label{sec:method_framework}

\PARbegin{DINOv2 (contrastive teacher--student training).}
DINOv2-style training~\cite{caron2020swav,caron2021dino,zhou2022ibot,oquab2023dinov2} compares student and teacher predictions across augmented image views.
For a teacher view $u$ and a student view $v$, the teacher produces features that are clustered online into a balanced prototype assignment $q(u)$ using Sinkhorn-Knopp normalization, while the student predicts an assignment $p_{\theta}(v)$.
Let $\mathcal{T}$ and $\mathcal{S}$ denote the teacher and student view sets.
The image-level loss is a cross-entropy over assignments,
\begin{equation}
    \mathcal{L}_{\mathrm{DINO}} =
    - \sum_{u \in \mathcal{T}} \sum_{\substack{v \in \mathcal{S}\\v \neq u}}
    q(u)^{\top}\log p_{\theta}(v).
\end{equation}
DINOv2 combines this view-level objective with iBOT-style patch prediction, where masked student patch predictions are matched to teacher patch assignments computed from the unmasked teacher view.
The teacher is an EMA model, and our implementation also uses the KoLeo regularizer~\cite{sablayrolles2018spreading} on image-level features.

\PAR{I-JEPA and V-JEPA (embedding prediction).}
JEPA-style training predicts representations of target regions from context representations~\cite{assran2023ijepa,bardes2024vjepa}.
Let $C$ be the visible context region and $\{B_k\}_{k=1}^{K}$ be target blocks.
The online encoder processes the context, $z_C=f_{\theta}(x_C)$, and a predictor $g_{\phi}$ predicts the representation of each target block.
An EMA target encoder $f_{\bar{\theta}}$ processes the unmasked image or clip and provides fixed target features,
\begin{equation}
    y_k = f_{\bar{\theta}}(x)_{B_k}, \qquad
    \hat{y}_k = g_{\phi}(z_C, B_k).
\end{equation}
Only the online encoder and predictor receive gradients.
We use an $\ell_1$ loss in representation space,
\begin{equation}
    \mathcal{L}_{\mathrm{JEPA}} =
    \frac{1}{K}\sum_{k=1}^{K}\|\hat{y}_k-y_k\|_1 .
\end{equation}
I-JEPA samples context and target blocks from one image, while V-JEPA uses spatiotemporal context and target regions from a video clip.

\PAR{I-MAE and V-MAE (masked reconstruction).}
MAE-style training masks a subset $M$ of tokens and reconstructs their RGB values from the visible tokens $V$~\cite{he2022mae,tong2022videomae,girdhar2023omnimae}.
The encoder processes only visible tokens, $z_V=f_{\theta}(x_V)$.
A decoder $d_{\psi}$ receives $z_V$ together with mask tokens and predicts the masked pixels as
\begin{equation}
    \hat{x}_M = d_{\psi}(z_V, M), \qquad
    \mathcal{L}_{\mathrm{MAE}} =
    \frac{1}{|M|}\sum_{i\in M}\|\hat{x}_i-x_i\|_2^2 .
\end{equation}
I-MAE masks image tokens, while V-MAE masks spatiotemporal tube tokens in a video clip.

\PAR{I-Diffusion and V-Diffusion (diffusion-style SSL).}
Diffusion-style SSL corrupts an image or video clip with Gaussian noise and trains a denoiser across noise levels~\cite{li2025jit}.
For a clean signal $x_0$, noise $\epsilon\sim\mathcal{N}(0,I)$, and clean weight $a_t$ at noise level $t$, the corrupted input in our JiT-style implementation is
\begin{equation}
    x_t = a_t x_0 + (1-a_t)\epsilon .
\end{equation}
The denoiser head predicts patch-space image or video values.
The training loss is applied in a velocity-normalized form,
\begin{equation}
    \mathcal{L}_{\mathrm{denoise}} =
    \mathbb{E}_{x_0,\epsilon,t}\left[
                                   \left\|
                                       \frac{h_{\theta}(x_t,t)-x_t}{1-a_t}
                                       -
                                       \frac{x_0-x_t}{1-a_t}
                                   \right\|_2^2
    \right].
\end{equation}
I-Diffusion applies this objective to images, while V-Diffusion applies it to video clips.

\subsection{Model Architecture}
\label{sec:method_arch}

\PARbegin{Why split image and temporal modeling?}
To study the complementarity of image and video SSL objectives, we need an architecture that can receive both image-level and video-level pretraining signals.
The image objective should train the visual representation directly, while the video objective should train the same visual representation in the context of motion and temporal consistency.
We therefore build all pretraining settings around a shared image-only encoder, and apply this encoder either to individual frames or framewise to video clips.
Video SSL objectives then add temporal processing after the shared encoder, so joint image+video pretraining can combine image losses on single frames with video losses on clips.
This design lets image SSL shape the per-frame visual backbone and lets video SSL add temporal supervision without replacing that backbone.

\PAR{Image-only encoder.}
All SSL methods use the same ViT image-only encoder architecture $E_{\theta}$.
For a given SSL method, this encoder is applied to images or framewise to all frames of a video clip with shared weights.
Some SSL objectives have a lightweight objective-specific head or decoder $G$ used only during pretraining and discarded during evaluation.
All attention layers in the encoder, temporal neck, and attention-based downstream heads use rotary positional encodings (RoPE)~\cite{su2024roformer}.

\begin{figure}[t]
    \centering
    \includegraphics[width=1.0\linewidth]{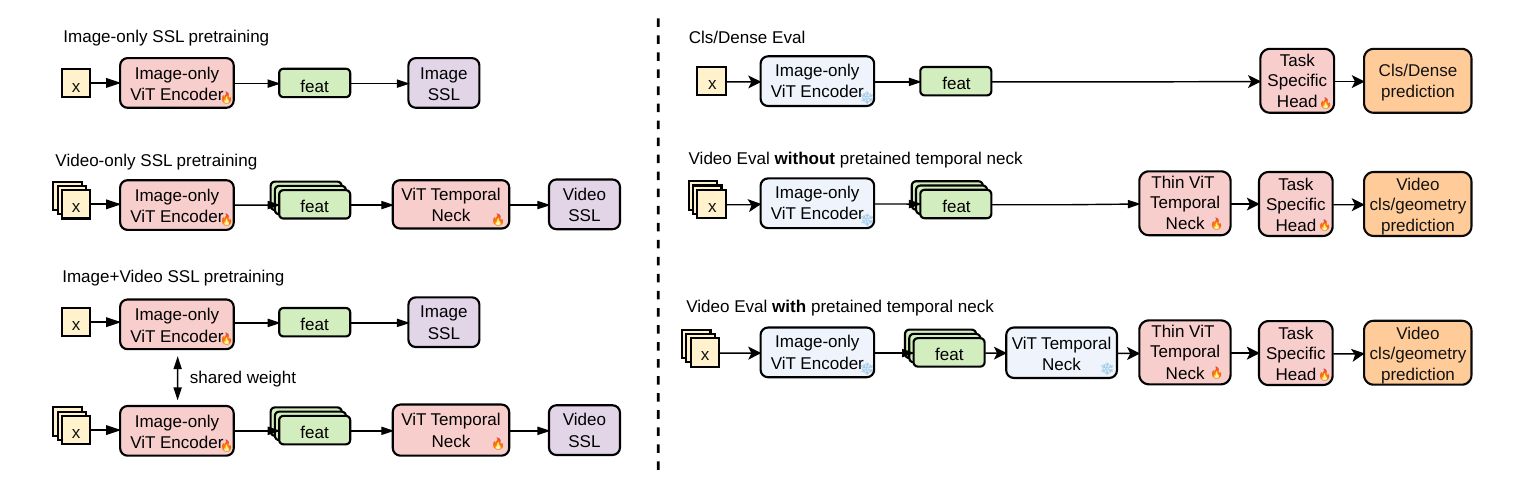}
    \caption{\textbf{Method architecture overview.}
    The image-only encoder is the common representation used for all image tasks, which matches image-task compute and parameter count across image and video SSL.
    Video objectives add a temporal neck after the framewise encoder; this neck can be reused or omitted during video downstream evaluation.}
    \label{fig:method_arch_overview}
\end{figure}

\PAR{Pretrained temporal neck.}
Video SSL objectives use the image-only encoder framewise and add a temporal neck $N_{\phi}$ that models interactions across frames.
After pretraining, we refer to this module as the \textbf{pretrained temporal neck}.
Figure~\ref{fig:method_arch_overview} shows the two architecture paths.
Image SSL trains the image-only encoder, while video SSL trains the same framewise image-only encoder followed by $N_{\phi}$.

For a video clip, the framewise encoder produces patch tokens $P \in \mathbb{R}^{B \times T \times N \times C}$ and CLS tokens $c \in \mathbb{R}^{B \times T \times C}$, where $B$ is batch size, $T$ is clip length, $N$ is the number of patch tokens per frame, and $C$ is the encoder width.
The temporal neck $N_{\phi}$ preserves this channel width and maps $(P,c)$ to tokens with the same shapes.
It adds temporal position embeddings to the per-frame CLS tokens, flattens the patch tokens over space and time, concatenates the $T$ CLS tokens with the $TN$ patch tokens, and applies a RoPE-based transformer stack over the resulting sequence.
The output sequence is then split back into patch tokens $\tilde{P} \in \mathbb{R}^{B \times T \times N \times C}$ and CLS tokens $\tilde{c} \in \mathbb{R}^{B \times T \times C}$.

\subsection{Controlled Pretraining}
\label{sec:method_variants}

We fix the source data, image-only encoder architecture, temporal-neck architecture when present, input resolution, video clip length, and optimizer-update count, as summarized in Table~\ref{tab:controlled_pretraining_setup}.
All SSL runs use K700 as the source data, with image SSL sampling individual frames from the K700 videos and video SSL sampling clips of consecutive frames.
The 80k-update budget is intentionally smaller than the official SSL schedules summarized in Table~\ref{tab:method_taxonomy}.
This limited pretraining budget lets us compare many SSL objectives with the same amount of compute.
We limit our claims to comparisons under this controlled budget, rather than to the best achievable performance of each objective. The base objective set is \{DINOv2~\cite{oquab2023dinov2}, I-JEPA~\cite{assran2023ijepa}, V-JEPA~\cite{bardes2024vjepa}, I-MAE~\cite{he2022mae}, V-MAE~\cite{tong2022videomae}, I-Diffusion~\cite{li2025jit}, V-Diffusion~\cite{li2025jit}\}.
For joint training, we focus on DINOv2-anchored combinations because DINOv2 gives the strongest semantic baseline among the evaluated single-objective methods.

The DINOv2+pixel-space SSL setting tests whether a pixel-level SSL objective can add fine spatial features for dense prediction while preserving DINOv2 semantics.
The DINOv2+V-MAE setting tests whether video SSL can add temporal or geometric structure to a DINOv2 encoder, and whether DINOv2 semantics can improve the image representation learned by video SSL.

Because the framewise image encoder and temporal neck are separate, these joint image and video SSL objectives use one multi-task training setup.
Image objectives attach to the image-encoder path, while video objectives attach to the framewise image encoder followed by the temporal neck.
In each training step, active objectives read their corresponding image or video SSL streams and their losses are summed.
Image objectives backpropagate through the shared image encoder only, whereas video objectives backpropagate through both the shared framewise image encoder and the temporal neck.
In joint image and video SSL runs, the image encoder receives gradients from all active SSL losses, while the temporal neck receives gradients only from video SSL losses.

\begin{table}[t]
    \caption{\textbf{Controlled pretraining setup.}
    The controlled comparisons keep source data, image-only encoder architecture, temporal-neck architecture when present, input resolution, video clip length, and optimizer-update count fixed.
    Training budget is matched by optimizer updates.}
    \label{tab:controlled_pretraining_setup}
    \centering
    \scriptsize
    \setlength{\tabcolsep}{3.6pt}
    \renewcommand{\arraystretch}{1.12}
    \begin{adjustbox}{max width=\linewidth}
        \begin{tabular}{@{}lccc@{}}
            \toprule
            & \multicolumn{1}{c}{This study}
            & \multicolumn{2}{c}{Representative original settings} \\
            \cmidrule(lr){2-2} \cmidrule(lr){3-4}
            Setting
            & Controlled runs
            & DINOv2~\cite{oquab2023dinov2}
            & V-MAE~\cite{tong2022videomae} \\
            \midrule
            Source data
            & K700
            & LVD-142M
            & SSV2 \\
            Frame resolution
            & $224^2$
            & $224^2 + 518^2$
            & $224^2$ \\
            Video clip length
            & 8 frames
            & --
            & \makecell{16 frames} \\
            Backbone
            & ViT-B/16
            & \makecell{ViT-g/14}
            & \makecell{ViT-B/16} \\
            Optimizer-update schedule
            & 80k
            & \makecell{625k + 10k}
            & 132k \\
            Training hardware
            & $8{\times}$ NVIDIA H100
            & \makecell{$256{\times}$ NVIDIA A100}
            & \makecell{$64{\times}$ NVIDIA V100} \\
            Total GPU-hours per run
            & 144--224 GPU-h
            & 22,016 GPU-h
            & 1,248 GPU-h \\
            \bottomrule
        \end{tabular}
    \end{adjustbox}
\end{table}

\subsection{Downstream Evaluation Tasks}
\label{sec:method_transfer}

\PARbegin{Frozen pretrained model.}
After pretraining, SSL objective-specific heads and decoders are removed, pretrained modules are frozen, and each downstream task trains only lightweight task-specific heads.
This makes the downstream tasks probe the learned representation through a comparable lightweight head.

\PAR{Image tasks.}
For classification, semantic segmentation, and monocular depth, we feed each image $x$ through the frozen image-only encoder $E_{\theta}$ and train a lightweight image-task head on top of the encoder features.
No pretrained temporal neck is used for these image tasks.
This keeps image-task parameter count and compute matched across image-only, video-only, and joint SSL pretraining.

\PAR{Video tasks.}
For each input clip, the frozen image-only encoder is applied framewise to produce patch and CLS tokens $(P,c)$.
We evaluate each video task in two variants using the same lightweight temporal head.
In the encoder-only variant, the downstream heads receive $(P,c)$ directly.
In the pretrained-neck variant, $(P,c)$ is first processed by the frozen pretrained temporal neck $N_{\phi}$, and the downstream heads receive $(\tilde{P},\tilde{c})$.
Because $N_{\phi}$ preserves token layout and feature dimension, the two variants differ in whether they reuse temporal pretraining.
Video classification and camera-pose heads use frame-level CLS features together with pooled patch features, while the tracking head reshapes patch tokens back to the spatial patch grid.

    \section{Experiments}
\label{sec:experiments_v2}

\newcommand{\expDINOvTwo}{DINOv2~\cite{oquab2023dinov2}}
\newcommand{\expIMAE}{I-MAE~\cite{he2022mae}}
\newcommand{\expVMAE}{V-MAE~\cite{tong2022videomae}}
\newcommand{\expIJEPA}{I-JEPA~\cite{assran2023ijepa}}
\newcommand{\expVJEPA}{V-JEPA~\cite{bardes2024vjepa}}
\newcommand{\expIDiffusion}{I-Diffusion~\cite{li2025jit}}
\newcommand{\expVDiffusion}{V-Diffusion~\cite{li2025jit}}
\newcommand{\w}{w/\ }
\newcommand{\wo}{w/o\ }

\subsection{Setup}
\label{sec:exp_v2_setup}

The experiments compare SSL objectives under the controlled pretraining setup from Sec.~\ref{sec:method_variants} and the frozen downstream evaluation from Sec.~\ref{sec:method_transfer}.
We keep the source dataset, image-only encoder architecture, temporal-neck architecture when present, input resolution, video clip length, and 80k optimizer-update schedule fixed while changing the SSL recipe.
This controls the main training configuration, but it does not equate FLOPs or wall-clock time because the objectives process different views or numbers of frames and use objective-specific modules.
All downstream tasks freeze the pretrained modules and train only lightweight task heads.

The evaluation covers image classification, semantic segmentation, monocular depth, video classification, point tracking, and relative camera pose.
Image tasks use only the frozen image-only encoder, even for models pretrained with video SSL.
Video tasks use the framewise image-only encoder and, for video SSL methods, can also reuse the frozen pretrained temporal neck.
This split keeps image comparisons matched and lets the video evaluation separate the shared encoder from temporal parameters learned during SSL pretraining.

The section is organized around four questions.
First, we compare prior SSL recipes under the shared data, encoder, and update schedule.
Second, we ask whether an image pixel objective can complement DINOv2 on dense tasks.
Third, we ask whether DINOv2 can broaden a video-specialized representation while retaining temporal and geometric structure.
Fourth, we isolate the effect of reusing the pretrained temporal neck in video downstream tasks.
Within each table column, the task data, head, and selection rule are fixed, while the pretrained setting, including its temporal neck when applicable, changes.

\subsection{Shared Data, Encoder, and Update Budget}
\label{sec:exp_v2_same_budget}

Table~\ref{tab:same_budget_image_v2} compares the individual image and video SSL objectives on image downstream tasks.
ImageNet-1K tests global semantic classification.
Pascal VOC, Cityscapes, and ADE20K test dense semantic prediction.
NYUv2 and KITTI test monocular depth estimation.

\begin{table}[t]
    \caption{\textbf{Controlled, resource-constrained study of image and video SSL objectives on image tasks.}
    Feature-space prediction transfers more broadly than pixel reconstruction or denoising under the short schedule, with DINOv2 and I-JEPA ranking first and second on every image evaluation.}
    \label{tab:same_budget_image_v2}
    \centering
    \scriptsize
    \setlength{\tabcolsep}{2.8pt}
    \renewcommand{\arraystretch}{1.12}
    \begin{adjustbox}{width=\linewidth}
        \begin{tabular}{l c cccccc}
            \toprule
            \multirow[b]{3}{*}{Method} & \multirow[b]{3}{*}{Modality}
            & \multicolumn{1}{c}{Img cls}
            & \multicolumn{3}{c}{Img seg}
            & \multicolumn{2}{c}{Img depth} \\
            \cmidrule(lr){3-3} \cmidrule(lr){4-6} \cmidrule(lr){7-8}
            & & \makecell{ImageNet-1K\\Top-1 (\%) $\uparrow$}
            & \makecell{Pascal VOC\\mIoU (\%) $\uparrow$}
            & \makecell{Cityscapes\\mIoU (\%) $\uparrow$}
            & \makecell{ADE20K\\mIoU (\%) $\uparrow$}
            & \makecell{NYUv2\\RMSE $\downarrow$}
            & \makecell{KITTI\\RMSE $\downarrow$} \\
            \midrule
            \expDINOvTwo & img & \bestcell{62.2} & \bestcell{54.7} & \bestcell{39.7} & \bestcell{25.0} & \bestcell{0.97} & \bestcell{5.91} \\
            \expIMAE & img & 42.7 & 28.5 & 25.1 & 10.2 & 1.20 & 7.05 \\
            \expVMAE & vid & 23.6 & 11.5 & 15.4 & 4.4 & 1.21 & 8.41 \\
            \expIJEPA & img & \secondcell{48.3} & \secondcell{38.1} & \secondcell{27.5} & \secondcell{14.3} & \secondcell{1.11} & \secondcell{6.47} \\
            \expVJEPA & vid & 39.8 & 21.4 & 20.6 & 6.8 & 1.23 & 7.89 \\
            \expIDiffusion & img & 25.0 & 10.6 & 17.6 & 3.7 & 1.26 & 9.54 \\
            \expVDiffusion & vid & 17.4 & 6.2 & 14.2 & 2.8 & 1.26 & 9.51 \\
            \bottomrule
        \end{tabular}
    \end{adjustbox}
\end{table}

\PAR{Feature-space objectives learn better under limited resources.}
In Table~\ref{tab:same_budget_image_v2}, DINOv2 ranks first in all six image evaluations, while I-JEPA ranks second in all six.
The margin of DINOv2 over I-JEPA is $+13.9$ points on ImageNet-1K, between $+10.7$ and $+16.6$ mIoU across the three segmentation datasets, and it has lower RMSEs with $-0.14$ and $-0.56$ differences on NYUv2 and KITTI.
DINOv2 supervises the encoder at two levels.
Moreover, across MAE, JEPA, and Diffusion, the image-based versions outperform their corresponding video-based versions in 16 of 18 image comparisons. Thus, the video objectives do not automatically yield a stronger image encoder under a short, resource-limited learning schedule.

Table~\ref{tab:same_budget_video_v2} uses the same objectives and learning schedules, but now video tasks are evaluated.
K700 and SSV2 test clip-level recognition.
MOVi-F tests point tracking.
RE10K tests relative camera pose.
For video SSL rows with a temporal neck, this table reports the pretrained-neck variant; Sec.~\ref{sec:exp_v2_temporal_neck} isolates that choice.

\begin{table}[t]
    \caption{\textbf{Evaluation of the image and video objectives of Table~\ref{tab:same_budget_image_v2} on video tasks.}
    The task changes the winner, with DINOv2 strongest on appearance-driven recognition and V-MAE strongest on motion and geometry.}
    \label{tab:same_budget_video_v2}
    \centering
    \scriptsize
    \setlength{\tabcolsep}{3.2pt}
    \renewcommand{\arraystretch}{1.12}
    \begin{adjustbox}{width=\linewidth}
        \begin{tabular}{l c c c c c c}
            \toprule
            \multirow[b]{4}{*}{Method} & \multirow[b]{4}{*}{Modality}
            & \multicolumn{2}{c}{Vid cls}
            & \multicolumn{2}{c}{Vid pose}
            & \multicolumn{1}{c}{Vid track} \\
            \cmidrule(lr){3-4} \cmidrule(lr){5-6} \cmidrule(lr){7-7}
            & & \makecell{K700\\Top-1 (\%) $\uparrow$}
            & \makecell{SSV2\\Top-1 (\%) $\uparrow$}
            & \makecell{RE10K\\Rot. acc.\\@0.1 (\%) $\uparrow$}
            & \makecell{RE10K\\Trans. acc.\\@5.0 (\%) $\uparrow$}
            & \makecell{MOVi-F\\Jaccard (\%) $\uparrow$} \\
            \midrule
            \expDINOvTwo & img & \bestcell{59.8} & \secondcell{49.5} & 45.2 & 11.1 & 66.0 \\
            \expIMAE & img & 53.4 & 47.8 & 86.2 & 25.1 & \secondcell{70.3} \\
            \expVMAE & vid & 52.8 & \bestcell{51.3} & \bestcell{90.7} & \bestcell{34.8} & \bestcell{71.4} \\
            \expIJEPA & img & \secondcell{55.2} & 48.4 & 39.0 & 10.8 & 66.4 \\
            \expVJEPA & vid & 47.5 & 45.4 & 83.8 & 24.5 & 64.9 \\
            \expIDiffusion & img & 40.4 & 36.7 & 84.7 & 25.1 & 67.9 \\
            \expVDiffusion & vid & 41.5 & 39.3 & \secondcell{88.6} & \secondcell{31.8} & 69.7 \\
            \bottomrule
        \end{tabular}
    \end{adjustbox}
\end{table}

\PAR{Video tasks expose complementary specialization.}
In Table~\ref{tab:same_budget_video_v2}, DINOv2 leads K700 at $59.8$, whereas V-MAE leads SSV2, both RE10K pose metrics, and MOVi-F, winning four of five video evaluations.
Relative to DINOv2, V-MAE trades $-7.0$ points on K700 for gains of $+1.8$ on SSV2, $+45.5$/$+23.7$ on rotation/translation accuracy, and $+5.4$ on tracking.
This advantage of video pretraining over its image counterpart is consistent on the geometric tasks.
Each video objective outperforms its image counterpart on both pose metrics, giving six gains in six comparisons.
This agrees with prior evidence that video-masked reconstruction is best distinguished from image SSL on spatial and temporal tasks rather than recognition alone~\cite{tong2022videomae,carreira2024scaling4d}.
These results do not support a single ranking across all tasks.
Under the common update schedule, DINOv2 provides the strongest semantic representation and V-MAE the strongest spatiotemporal representation.
This complementarity motivates the use of DINOv2 as the base objective for our joint image+video SSL experiments in Sec.~\ref{sec:exp_v2_dinov2_video_ssl}.

\FloatBarrier

\subsection{Combining SSL Objectives for Dense Image Tasks}
\label{sec:exp_v2_pixel_losses}
The experiment reported in Table~\ref{tab:dinov2_dense_losses_v2} asks whether adding an image-pixel objective to the DINOv2 objectives improves dense image evaluation.
DINOv2 appears alone and with either I-MAE reconstruction or I-Diffusion denoising added as an extra objective.
ImageNet-1K is included as a semantic check, while the segmentation and depth columns are the main dense-task evaluations.

\begin{table}[t]
    \caption{\textbf{Adding pixel-space SSL objectives to DINOv2.}
    Under the shared schedule, masked reconstruction complements DINOv2 by improving classification and every segmentation evaluation.}
    \label{tab:dinov2_dense_losses_v2}
    \centering
    \scriptsize
    \setlength{\tabcolsep}{3.2pt}
    \renewcommand{\arraystretch}{1.12}
    \begin{adjustbox}{width=\linewidth}
        \begin{tabular}{l cccccc}
            \toprule
            \multirow[b]{3}{*}{Method}
            & \multicolumn{1}{c}{Img cls}
            & \multicolumn{3}{c}{Img seg}
            & \multicolumn{2}{c}{Img depth} \\
            \cmidrule(lr){2-2} \cmidrule(lr){3-5} \cmidrule(lr){6-7}
            & \makecell{ImageNet-1K\\Top-1 (\%) $\uparrow$}
            & \makecell{Pascal VOC\\mIoU (\%) $\uparrow$}
            & \makecell{Cityscapes\\mIoU (\%) $\uparrow$}
            & \makecell{ADE20K\\mIoU (\%) $\uparrow$}
            & \makecell{NYUv2\\RMSE $\downarrow$}
            & \makecell{KITTI\\RMSE $\downarrow$} \\
            \midrule
            \expDINOvTwo & \secondcell{62.2} & \secondcell{54.7} & 39.7 & \secondcell{25.0} & \secondcell{0.97} & \secondcell{5.91} \\
            + \expIMAE & \bestcell{62.8} & \bestcell{55.4} & \bestcell{40.2} & \bestcell{25.4} & \bestcell{0.95} & 5.97 \\
            + \expIDiffusion & 60.4 & 48.9 & \secondcell{39.8} & \secondcell{25.0} & 0.98 & \bestcell{5.88} \\
            \bottomrule
        \end{tabular}
    \end{adjustbox}
\end{table}

\PAR{Masked reconstruction complements DINOv2.}
Adding I-MAE improves five of the six point estimates.
ImageNet-1K rises by $+0.6$ points, all three segmentation datasets rise by between $+0.4$ and $+0.7$ mIoU, and NYUv2 RMSE falls by $-0.02$.
Only KITTI changes adversely, by $+0.06$ RMSE.
These gains are modest, but their breadth is notable, as the added raw-pixel target improves every segmentation dataset, even though DINOv2 already contains masked patch supervision.
Adding I-Diffusion mostly harms transfer under the resource-limited schedule.
It decreases performance on ImageNet-1K, Pascal VOC, and NYUv2, leaves ADE20K unchanged, and produces only small gains on Cityscapes and KITTI.
The results seem to indicate that objective compatibility matters more than the mere presence of extra pixel supervision.
Within 80k updates, I-MAE is a low-trade-off complement to DINOv2, whereas simply adding a second pixel-space loss like I-Diffusion is insufficient.

\FloatBarrier

\subsection{Adding DINOv2 to Video MAE}
\label{sec:exp_v2_dinov2_video_ssl}

The experiment reported in Table~\ref{tab:dinov2_video_ssl_v2} asks whether adding DINOv2 can broaden the representation learned by video SSL, as the experiments reported in Sec.~\ref{sec:exp_v2_same_budget} demonstrated the complementarity of the DINOv2 and V-MAE objectives.
This experiment directly compares standalone DINOv2, standalone V-MAE, and their combination.
Image classification and semantic segmentation assess whether the shared image encoder captures stronger semantic features, while video classification assesses clip-level semantics.
We evaluate point tracking on MOVi-F and relative camera pose on RE10K.

\begin{table}[t]
    \caption{\textbf{Adding DINOv2 to V-MAE.}
    Joint training closes much of the semantic--geometric gap, improving eight of nine DINOv2 evaluations while V-MAE retains the strongest pose and tracking.}
    \label{tab:dinov2_video_ssl_v2}
    \centering
    \scriptsize
    \setlength{\tabcolsep}{1.4pt}
    \renewcommand{\arraystretch}{1.12}
    \newcommand{\dinovideohead}[1]{\scalebox{0.7}{\makecell{#1}}}
    \begin{adjustbox}{width=\linewidth}
        \begin{tabular}{l c c c c c c c c c}
            \toprule
            \multirow[b]{3}{*}{Method}
            & \multicolumn{1}{c}{Img cls}
            & \multicolumn{3}{c}{Img seg}
            & \multicolumn{2}{c}{Vid cls}
            & \multicolumn{2}{c}{Vid pose}
            & \multicolumn{1}{c}{Vid track} \\
            \cmidrule(lr){2-2} \cmidrule(lr){3-5} \cmidrule(lr){6-7} \cmidrule(lr){8-9} \cmidrule(lr){10-10}
            & \dinovideohead{ImageNet-1K\\Top-1 (\%) $\uparrow$}
            & \dinovideohead{Pascal VOC\\mIoU (\%) $\uparrow$}
            & \dinovideohead{Cityscapes\\mIoU (\%) $\uparrow$}
            & \dinovideohead{ADE20K\\mIoU (\%) $\uparrow$}
            & \dinovideohead{K700\\Top-1 (\%) $\uparrow$}
            & \dinovideohead{SSV2\\Top-1 (\%) $\uparrow$}
            & \dinovideohead{RE10K\\Rot. acc.\\@0.1 (\%) $\uparrow$}
            & \dinovideohead{RE10K\\Trans. acc.\\@5.0 (\%) $\uparrow$}
            & \dinovideohead{MOVi-F\\Jaccard (\%) $\uparrow$} \\
            \midrule
            \expDINOvTwo & \secondcell{62.2} & \secondcell{54.7} & \secondcell{39.7} & \secondcell{25.0} & \secondcell{59.8} & 49.5 & 45.2 & 11.1 & \secondcell{66.0} \\
            \expVMAE & 23.6 & 11.5 & 15.4 & 4.4 & 52.8 & \secondcell{51.3} & \bestcell{90.7} & \bestcell{34.8} & \bestcell{71.4} \\
            V-MAE + DINOv2 & \bestcell{62.7} & \bestcell{55.0} & \bestcell{41.1} & \bestcell{25.7} & \bestcell{60.8} & \bestcell{52.8} & \secondcell{81.6} & \secondcell{21.8} & 65.7 \\
            \bottomrule
        \end{tabular}
    \end{adjustbox}
\end{table}

\PAR{DINOv2+V-MAE turns two specialists into one model for image semantics and video geometry.}
The joint use of DINOv2 and video MAE outperforms both standalone trainings across all six classification and segmentation evaluations.
This includes ImageNet-1K, all three segmentation datasets, and the video classification datasets K700 and SSV2.
Against standalone DINOv2 in Tables~\ref{tab:same_budget_image_v2} and~\ref{tab:same_budget_video_v2}, the exact changes are $+0.5$ on ImageNet-1K, $+0.3$ on Pascal VOC, $+1.4$ on Cityscapes, $+0.7$ on ADE20K, $+1.0$ on K700, $+3.3$ on SSV2, $+36.4$/$+10.7$ on rotation/translation accuracy, and $-0.3$ on MOVi-F.
The joint recipe therefore improves eight of the nine evaluations relative to DINOv2.
Relative to standalone V-MAE, these semantic gains come with lower pose and tracking performance; compared with DINOv2, however, the joint recipe retains near-identical tracking performance while adding substantially stronger pose performance.
Under the constrained update schedule and shared joint recipe, DINOv2+V-MAE combines the semantic and spatiotemporal specialists into a single frozen representation that performs strongly across image semantics, video recognition, and pose.
\FloatBarrier

\subsection{What Structure Is Learned Where?}
\label{sec:exp_v2_temporal_neck}
The previous experiment showed that image and video SSL losses can be complementary, and joint training with both types of objectives can improve the effectiveness of SSL in resource-limited settings.
However, Tab.~\ref{tab:dinov2_video_ssl_v2} also showed that performance on semantic tasks increased, whereas performance on geometric-understanding tasks decreased, compared to the best single-objective training.
Ideally, we would want to prevent this decrease, and to do so in future work, we aim to first better understand which types of information (semantic and geometric) are learned where in the model (the image encoder or the video temporal neck).
For this, we conduct an experiment evaluating with and without the pretrained temporal neck. By evaluating the relative decrease in performance on semantic and geometric tasks, we gain insight into which structures are learned where.
All evaluation experiments use a lightweight temporal neck whose weights are not pretrained and are learned solely from the labels of the evaluation datasets; see also Fig.~\ref{fig:method_arch_overview}.\newline

\begin{table}[t]
    \caption{\textbf{Reusing the pretrained temporal neck.}
    For image-and-video pretraining, the pretrained temporal neck is mainly crucial for video geometric tasks.}
    \label{tab:temporal_neck_effect_revised_v2}
    \centering
    \scriptsize
    \setlength{\tabcolsep}{1.2pt}
    \renewcommand{\arraystretch}{1.08}
    \newcommand{\temporalneckhead}[1]{\scalebox{0.8}{\makecell{#1}}}
    \newcommand{\temporalneckval}[1]{\scalebox{0.82}{#1}}
    \newcommand{\temporalnecksign}[1]{\scalebox{0.75}{#1}}
    \begin{tabular*}{\linewidth}{
        @{\extracolsep{\fill}}
        l
        c@{\hspace{3pt}}c
        c@{\hspace{3pt}}c
        c@{\hspace{3pt}}c
        c@{\hspace{3pt}}c
        c@{\hspace{3pt}}c
        @{}
    }
            \toprule
            \multirow[b]{3}{*}{Method}
            & \multicolumn{4}{c}{Vid cls}
            & \multicolumn{4}{c}{Vid pose}
            & \multicolumn{2}{c}{Vid track} \\
            \cmidrule(lr){2-5} \cmidrule(lr){6-9} \cmidrule(lr){10-11}
            & \multicolumn{2}{c}{\temporalneckhead{K700\\Top-1 (\%) $\uparrow$}}
            & \multicolumn{2}{c}{\temporalneckhead{SSV2\\Top-1 (\%) $\uparrow$}}
            & \multicolumn{2}{c}{\temporalneckhead{RE10K\\Rot. acc.\\@0.1 (\%) $\uparrow$}}
            & \multicolumn{2}{c}{\temporalneckhead{RE10K\\Trans. acc.\\@5.0 (\%) $\uparrow$}}
            & \multicolumn{2}{c}{\temporalneckhead{MOVi-F\\Jaccard (\%) $\uparrow$}} \\
            \cmidrule(lr){2-3} \cmidrule(lr){4-5} \cmidrule(lr){6-7}
            \cmidrule(lr){8-9} \cmidrule(lr){10-11}
            \multicolumn{1}{l}{\temporalnecksign{\textit{Pretrained neck}}}
            & \temporalnecksign{$\xmark$} & \temporalnecksign{$\checkmark$}
            & \temporalnecksign{$\xmark$} & \temporalnecksign{$\checkmark$}
            & \temporalnecksign{$\xmark$} & \temporalnecksign{$\checkmark$}
            & \temporalnecksign{$\xmark$} & \temporalnecksign{$\checkmark$}
            & \temporalnecksign{$\xmark$} & \temporalnecksign{$\checkmark$} \\
            \midrule
            \expVMAE & \temporalneckval{43.3} & \improvecell{\temporalneckval{52.8}} & \temporalneckval{39.1} & \improvecell{\temporalneckval{51.3}} & \temporalneckval{84.4} & \improvecell{\temporalneckval{90.7}} & \temporalneckval{24.9} & \improvecell{\temporalneckval{34.8}} & \temporalneckval{69.2} & \improvecell{\temporalneckval{71.4}} \\
            V-MAE + DINOv2 & \temporalneckval{60.0} & \improvecell{\temporalneckval{60.8}} & \temporalneckval{49.2} & \improvecell{\temporalneckval{52.8}} & \temporalneckval{53.3} & \improvecell{\temporalneckval{81.6}} & \temporalneckval{11.5} & \improvecell{\temporalneckval{21.8}} & \temporalneckval{66.9} & \degradecell{\temporalneckval{65.7}} \\
            \bottomrule
    \end{tabular*}
\end{table}

\PAR{The temporal neck learns geometric-temporal structure.}
Table~\ref{tab:temporal_neck_effect_revised_v2} evaluates video SSL methods with and without the frozen pretrained temporal neck.
For V-MAE, we see that important semantic and geometric structure is learned in the temporal neck, as removing it consistently reduces performance.
However, when including the DINOv2 objectives with V-MAE, we see that removing the temporal neck primarily hurts the geometric tasks, while the semantic tasks largely maintain their accuracy.
This shows that in this combined image-and-video training, the image encoder can learn strong image- and video-level semantic structure, while the temporal neck remains important for geometric structure.
This interesting insight suggests that improved joint image-and-video training strategies might be possible by better targeting which objectives act on which parts of the architecture.

    \section{Conclusion}
We presented a controlled study of image and video self-supervised learning objectives under a fixed and limited pretraining budget. By matching data, architecture, and compute across methods, we isolated the impact of the SSL objective on representation quality and transfer performance. Our results show that DINOv2 provides the strongest overall performance in this resource-constrained setting, while joint image-video pretraining with VideoMAE further improves semantic image and video understanding. At the same time, these gains come with reduced performance on tracking and camera-pose estimation, highlighting a tradeoff between semantic and geometric representation learning.

More broadly, our findings align with the goal of learning effective visual representations under limited resources: \textbf{careful objective selection can have a large impact for resource-limited pretraining, and complementary image and video supervision can improve representation quality with negligible impact on the overall training budget}. We hope this benchmark and analysis help guide the development of future resource-efficient visual foundation models that better balance semantic, temporal, and geometric understanding. For future work, our study indicates that a deeper investigation into how best to combine DINOv2 objectives in video pretraining to improve overall image and video understanding is a promising direction to obtain effective VFMs under resource limitations.

    \newpage
\section*{Acknowledgements}
The project EdgeAI “Edge AI Technologies for Optimised Performance Embedded Processing” is supported by the Chips Joint Undertaking and its members including top-up funding by Austria, Belgium, France, Greece, Italy, Latvia, Netherlands, and Norway under grant agreement No 101097300.

The project Ecomobility has received funding within the Chips Joint Undertaking (Chips JU), a public-private partnership in collaboration with the Horizon Europe (HORIZON) Framework Programme and National Authorities under grant agreement number 101112306.

This work made use of the Dutch national e-infrastructure with the support of the SURF Cooperative, using grant nos. EINF-16405, EINF-19284 and EINF-12907, which is financed by the Dutch Research Council (NWO).
The authors acknowledge the Supercomputing Center of the Eindhoven University of Technology~\cite{tue_spike_cluster} for providing access to the SPIKE-1 supercomputer and assistance with the various computing resources available.

\section*{Disclaimer}
Funded by the European Union.
Views and opinions expressed are, however, those of the author(s) only and do not necessarily reflect those of the European Union or the Chips Joint Undertaking.
Neither the European Union nor the granting authority can be held responsible for them.

    \bibliographystyle{splncs04}
    \bibliography{main}

\end{document}